\documentclass[11pt,a4paper]{article}
\usepackage[hyperref]{emnlp2018}
\usepackage{times}
\usepackage{latexsym}
\usepackage{graphicx}

\usepackage{url}

\aclfinalcopy 

\title{Super Characters: A Conversion from Sentiment Classification to Image Classification}

\author{Baohua Sun, Lin Yang, Patrick Dong, Wenhan Zhang, Jason Dong, Charles Young\\
Gyrfalcon Technology Inc.\\
1900 McCarthy Blvd. Milpitas, CA 95035\\
{\tt\small \{baohua.sun,lin.yang,patrick.dong,wenhan.zhang\}@gyrfalcontech.com}}
\date{}

\begin{document}
\maketitle
\begin{abstract}
We propose a method named Super Characters for sentiment classification. This method converts the sentiment classification problem into image classification problem by projecting texts into images and then applying CNN models for classification. Text features are extracted automatically from the generated Super Characters images, hence there is no need of any explicit step of embedding the words or characters into numerical vector representations. Experimental results on large social media corpus show that the Super Characters method consistently outperforms other methods for sentiment classification and topic classification tasks on ten large social media datasets of millions of contents in four different languages, including Chinese, Japanese, Korean and English. 

\end{abstract}

\section{Introduction}
Sentiment classification is an interesting topic that has been studied for many years~\cite{hatzivassiloglou1997predicting, pang2002thumbs, hong2015sentiment}. Word embedding is a widely used technique for sentiment classification tasks, which embeds the words into numerical vector representation before the sentences are fed into models for classification ~\cite{mikolov2013distributed, le2014distributed, pennington2014glove, yu2017joint, cao-lu-zhou-li:2018:AAAI2018}. For sequential input, RNNs are usually used and have very good results for text classification tasks~\cite{lai2015recurrent,tang2015document}. Recently, there are also works using Convolutional Neural Networks (CNN) for text classification~\cite{kim2014convolutional,severyn2015twitter,vaswani2017attention, bai2018empirical}. CNN models have feature extraction and classification in a whole model, which require no need of manually extracting features from images and are proved to be successful in image classification tasks~\cite{lecun1998gradient,krizhevsky2012imagenet,simonyan2014very,szegedy2015going,he2016deep,hu2017squeeze}. There are also works on character level text classifications~\cite{ zhang2015character, zhang2015text,kim2016character}. However, the input for CNNs are still using the embedding vectors.

\citet{zhang2017encoding} had studied the different ways of encoding Chinese, Japanese, Korean (CJK) and English languages for text classification. These encoding mechanisms include Onehot encoding, embedding and images of character glyphs. Comparisons with linear models, fastText~\cite{joulin2016bag}, and convolutional networks were provided. This work studied 473 models, using 14 large-scale text classification datasets in 4 languages including Chinese, English, Japanese and Korean. 

Our work in this paper is based on the datasets provided in~\cite{zhang2017encoding} and downloadable at ~\cite{Lecun14Datasets2017}. Different from existing methods, our method has no explicit step of embedding the text into numerical vector representations. Instead, we project the text into images and then directly feed the images into CNN models to classify the sentiments.

Before introducing the details of our solution, let us first look at how humans read text and do sentiment analysis. Humans read sentences and can immediately understand the sentiment of the text; Humans can also read multiple lines at a first sight of paragraphs and get the general idea instantly. This fast process consists of two steps. First, the texts are perceived by human's eyes as a whole picture of text, while the details of this picture are block-built by many characters. Second, the image containing the texts are fed into the brain. And then the human brain processes the image of texts to output the sentiment classification results. During the processing, the human brain may recognize words and phrases as the intermediate results, in order to further analyze the sentiment. However, if we treat the human brain as a black box system, its input is the image of texts received by the eyes, and its output is the sentiment classification result.

In this paper, we propose a two-step method that is similar to how humans do sentiment classification. We tested our method using the datasets provided by~\citet{zhang2017encoding} on text classification tasks for social media contents from different countries in four languages, including English, Chinese, Japanese, and Korean. And compared with other existing methods, including fastText, EmbedNet, OnehotNet, and linear models. The results show our method consistently outperforms other method on these datasets for sentiment classification tasks.

\section{Super Characters Method}
\begin{figure}[t]
\begin{center}
   \includegraphics[width=0.45\linewidth]{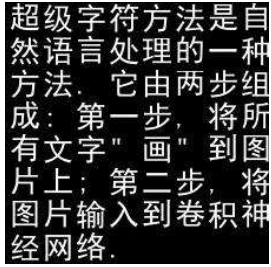}
\end{center}
   \caption{A Super Character example.}
\label{SuperCharacter}
\end{figure}

The Super Characters method converts the sentiment classification problem into an image classification problem. It is defined in two steps.
\begin{itemize}
\item First, the texts, e.g. sentences or paragraphs, are ``drawn" onto blank images, character by character. For example, a generated Super Characters image from Chinese text inputs (including punctuation marks) is shown in Figure~\ref{SuperCharacter}. The Chinese text means ``Super characters are a method for NLP. It consists of two steps: Frist, ``draw" text onto images; second, feed images into CNN". Each generated Super Character image is attached with the same sentiment labels as its original text.
\item Second, feed the generated Super Characters images with its labels to train CNN models.
\end{itemize}

The information embedded in the Super Characters image is near identical to that in the original text, so we convert the sentiment classification problem into image classification problem. The Super Characters images are similar to how humans perceive text: as whole pictures containing text, whether printed on paper, projected on a screen, or written by hand. After the texts are converted to images, the performance of our text classification method is determined by the accuracy of image classification models. For large scale image classification tasks, CNN models such as ResNet\cite{ResNet} have outperformed humans in image classification tasks as an end to end solution. Thus, if we feed the Super Characters images into CNN models such as ResNet, we expect the text classification using this 2-step pipeline to have a high accuracy.

For detailed implementation of projecting text into Super Characters image, there are a few settings to configure, including the image size of the whole Super Characters image; number of characters per row/column; size of each character; cut-length, which is the length of sentence to cut/padding in order to fit into the image; the fonts used to project each character into an image, and so on. 

For Latin languages, we have the option of projecting text at the word level or at the alphabet level, which will make differences at some cases. For example, how to handle line-change if a word is at the end of a row in the Super Characters image and can't fit in the residual space in that row. If we separate the words into separate alphabets, we can fit as many characters in the residual space in that row and change to the next row for the rest of the alphabets in that word. Or, if we keep the word as a whole entity and avoid spliting, we have to change to the next line for that word. 

For example, here are the settings used in one of our experiments in Section 3. We use a fixed image size of 224x224. And we also prefer integer numbers of characters in each row and having same-sized characters. Thus, we prefer to use 8x8=64, or 28x28=784, or 32x32=1024 characters per image. And we set the cut/padding length as the same. That also means, for 8x8 =64 settings, we will have 8 characters per row, and we have 8 rows in total. And each character is set to be of size 224/8=28 square pixels. The Arial\_Unicode\_MS font is selected as font to draw text onto image. For padding, we just draw nothing on the image. 

From the definition and description of Super Character, we can see it has the following advantages.
1. Its speed is not sensitive to the length of the text input, so it can easily handle long and short texts input. This advantage will be more obvious when the input text is long, because super character using CNN as model will be parallel processing the input. And the processing time is invariant for  training and inference. 
2. The feature engineering work is no longer needed, which includes generating manmade features of each character related to the culture behind each language. The Super Characters image is treated as an input for CNN models, and the feature extraction task are handled automatically by CNN models.
3. Similar to image classification using CNN networks which requires large amount of labeled image data, this method of Super Characters for sentiment analysis also requires large amount of labeled text data.

\section{Experiments}
\label{sec:Experiments}

\subsection{Sentiment Classification on Large Datasets from Online Social Media}
Ten of 14 datasets provided by~\cite{zhang2017encoding} were tested on, a brief description of which is provided:

\textbf{Dianping}: Chinese restaurant reviews were evenly split as follows: 4 and 5 star reviews were assigned to the positive class while 1-3 star reviews were in the negative class.

\textbf{JD Full}: Chinese shopping reviews wer evenly split for predicting full five stars. \textbf{JD Binary} Chinese shopping reviews are evenly split into positive (4-and-5 star reviews) and negative (1-and-2 star reviews) sentiments, ignoring 3-star reviews.

\textbf{Rakuten Full} Japanese shopping reviews were evenly split into predicting full five stars. \textbf{Rakuten Binary} Japanese shopping reviews were evenly split into positive (4-and-5 star reviews) and negative (1-and-2 star reviews) sentiments, removing duplicates and ignoring 3-star reviews.

\textbf{11st Full} Korean shopping reviews were evenly split into predicting full five stars. \textbf{11st Binary} Korean shopping reviews were evenly split into identifying positive (4-and-5 star reviews) or negative (1-3 star reviews) sentiments.

\textbf{Amazon Full}: English shopping reviews were evenly split into predicting full five stars.

\textbf{Ifeng}: First paragraphs of Chinese news articles from 2006-2016 were evenly split into 5 news channels.

\textbf{Chinanews}: Chinese news articles from 2008-2016 were evenly split into 7 news channels, removing duplicates.

The statistics of these datasets are given in Table \ref{table:DataSetsStatistics}. We can see that eight out of the ten datasets has more than millions of samples in training, and the largest datasets have 4 millions of samples in training set. The test datasets are in the range of 1/4 to 1/16 of the training datasets respectively. The languages used in these datasets include Chinese, Japanese, Korean, and English. And the number of classes ranges from 2, 5 to 7. 

For each dataset, we generate Super Characters images first. We draw text with the Python Imaging Library (PIL)~\cite{PIL}, and set all the Super Character image sizes to 224x224 pixels, the background set to black. For long text inputs such as paragraphs or articles, the length of which is different so we set a cut-length from the beginning of the news article. Although we may forcely cut the input and ignore the rest, this cut-length still works well since the first few sentences usually convey the general information about the whole contents. For other text sources and tasks, the starting point of the text for the cut-length may change accordingly. For each experiment, we determine the estimated cut-length by using a threshold on sentence lengths. We have only tried one cut-length of 14x14=196 for every experimental data set. We set the size of each character as 224/14=16 square pixels. 

And then, we feed the generated Super Characters to train CNN models. We use successful pretrained model SENet-154~\cite{hu2017squeeze} in the ImageNet competition~\cite{ImageNet}, which is the winner in ImageNet2017 competition and achieves 81.32\% Top1 accuracy and 95.53\% Top5 accuracy. We used pretrained model downloadable at \cite{SENet154} because it gave a good initialization for transfer learning tasks. We changed the last layer to the corresponding number of categories in each data set to train on the Super Characters images.

The sentiment classification results on test datasets are shown in Table \ref{table:LeCunDatasetResults}. The accuracy numbers for the models of OnehotNet, EmbedNet, Linear Model, and fastText are given by \cite{zhang2017encoding}. Note that in \cite{zhang2017encoding}, each model is tried with different encoding methods. For example, OnehotNet uses 4 different encodings, EmbedNet uses 10, Linear Models uses 11, and fastText uses 10. We only listed the best results for each existing method across different encodings. And compare our results with the best of them. That means we compare our results with the finetuned best encoding of each existing model in \ref{table:LeCunDatasetResults}. From the results we can see that our Super Characters method (short as S.C.) consistently outperforms other methods, even with their best encodings. 

\begin{table*}[t!]
\begin{center}
\begin{tabular}{|c|c|c|c|c|c|}
\hline Dataset&Short Name& Language&Classes&Train&Test\\ \hline
Dianping&D.P.  & Chinese&2&2,000,000&500,000\\ 
JD full&JD.f&Chinese&5&3,000,000&250,000 \\
JD binary&JD.b&Chinese&2&4,000,000&360,000 \\
Rakuten full&RKT.f&Japanese&5&4,000,000&500,000 \\
Rakuten binary&RKT.b&Japanese&2&3,400,000&400,000 \\
11st full&11st.f&Korean&5&750,000&100,000 \\
11st binary&11st.b&Korean&2&4,000,000&400,000 \\
Amazon full&AMZ.f&English&5&3,000,000&650,000 \\
Ifeng&Ifeng&Chinese&5&800,000&50,000 \\
Chinanews&Cnews&Chinese&7&1,400,000&112,000 \\
\hline
\end{tabular}
\end{center}
\caption{\label{table:DataSetsStatistics} Datasets statistics used in Table \ref{table:LeCunDatasetResults} and short names used for convenience.}
\end{table*}

\begin{table*}[t!]
\begin{center}
\begin{tabular}{|c|c|c|c|c|c|c|c|c|c|c|}
\hline  Model &  D.P. & JD.f & JD.b&RKT.f&RKT.b&11st.f&11st.b&AMZ.f&Ifeng&Cnews\\ \hline
OnehotNet & 76.83&51.90&90.69&54.90&94.07&67.57&86.70&57.79&83.51&89.38 \\ 
EmbedNet &  76.40 &51.71&90.81&54.80&93.93&67.71&86.75&56.30&82.99&89.45\\
Linear & 76.97&51.82&91.18&54.74&93.37&56.58&86.60&57.30&81.70&89.24 \\
fastText & 77.66&52.01&91.28&56.73&94.55&61.42&86.89&59.98&83.69&90.90 \\ \hline
S.C.(ours) & \bf 77.80&\bf54.10&\bf92.20&\bf57.70&\bf94.85&\bf68.70&\bf87.60&\bf60.70&\bf84.40&\bf92.00 \\
\hline
\end{tabular}
\end{center}
\caption{\label{table:LeCunDatasetResults} Results of our Super Character (SC) method against other models on datasets provided by~\cite{zhang2017encoding}.}
\end{table*}

\subsection{Experiments on THUCTC corpus}
THUCTC~\cite{THUCTC} was provided by the Tsinghua University NLP lab in 2016. It totals 836075 documents after downloaded, covering 14 topics including 24373 Game, 37098 Finance, 63086 Politics, 50849 Society, 32586 Living, 20050 Real Estate, 7588 Lottery, 92632 Entertainment, 41936 Education, 13368 Fashion, 3578 Constellation, 162929 Technology, 131604 Sports and 154398 Stocks. The majority of the documents are long articles with hundreds or sometimes thousands of characters in multiple sentences or paragraphs. We use a cut-length of 28x28=784, each having an 8x8 pixel size and utilize simhei font for Super Characters on the THUCTC data. In Table~\ref{table:THUCTC_results}, we showed our Super Character method using ResNet-50 (SC+ResNet50) attained an accuracy of 94.85\% and our Super Character method using ResNet-152 (SC+ResNet152) attained an accuracy of 94.35\%, while the result given by Sun et al. (2016) achieved only an accuracy of 88.6\% using LibLinear. LibLinear~\cite{fan2008liblinear} implements linear SVMs and logistic regression models trained using a coordinate descent algorithm. Our models reduce the error by 50.4\% compared to this existing model.
\begin{table}[t!]
\begin{center}
\begin{tabular}{|l|r|}
\hline \bf Model & \bf Accuracy \\ \hline
LibLinear \cite{THUCTC} & 88.6\% \\ \hline
SC+ResNet-50 (ours) & \bf 94.85\% \\
SC+ResNet-152 (ours) & 94.35\% \\
\hline
\end{tabular}
\end{center}
\caption{\label{table:THUCTC_results} Results of our Super Character (SC) method against other models on THUCTC data set.}
\end{table}
\subsection{Experiments on Fudan Corpus}
The Fudan corpus~\cite{FudanCorpus} contains 9804 documents of long sentences and paragraphs in 20 categories. We use the same split as ~\cite{xu2016improve, cao-lu-zhou-li:2018:AAAI2018} in selecting the same 5 categories: 1218 environmental, 1022 agricultural, 1601 economical, 1025 political and 1254 sport documents; 70\% of the total data is used for training and the rest for testing. 
\begin{itemize}
\item SC+ResNet-50: Using a ResNet-50 model pretrained on the ImageNet dataset, we fine-tuned the transfer learning model on the new generated super character dataset.
\item SC+ResNet-50-THUCTC: Using a ResNet-50 model pretrained on THUCTC data, we fine-tuned the trasfere learning model on the new generated super character dataset.
\end{itemize}
We used a cut-length of 28x28=784 and words of pixel size 8x8 with the simhei font for our Super Characters in this experiment. In Table~\ref{table:fudan_results}, the first 7 rows of model accuracies for different algorithms are given by~\cite{cao-lu-zhou-li:2018:AAAI2018}. We can see that our SC+ResNet-50-THUCTC model attained an accuracy of 97.8\% while the best existing method achieved only a 95.3\% accuracy. Our SC+ResNet50-THUCTC model reduces the error by 53.2\% compared with the best existing model. The SC+ResNet-50 model with 95.7\% accuracy also outperforms the best existing model. The pretrained model on THUCTC dataset gives 2.1\% accuracy improvement than SC+ResNet-50 model, which means pretrained models on the same language and a larger dataset will help for a better initialization and better model. For this data set, we did not delete the non-Chinese characters as (Cao et al., 2018) did. The result shows that our simple projection from text to Super Characters image is easy to implement and very robust. Users do not even need to perform complicated preprocessing techniques for the data.
\begin{table}[t!]
\begin{center}
\begin{tabular}{|l|r|}
\hline \bf Model & \bf Accuracy \\ \hline
skip-gram \cite{mikolov2013distributed} & 93.4\% \\ 
cbow \cite{mikolov2013distributed} & 93.4\% \\ 
GloVe \cite{pennington2014glove} & 94.2\% \\ 
CWE \cite{chen2015does} & 93.2\% \\ 
GWE \cite{su2017learning} & 94.3\% \\ 
JWE \cite{yu2017joint} & 94.2\% \\ 
cw2vec \cite{cao-lu-zhou-li:2018:AAAI2018} & 95.3\% \\ \hline
SC+ResNet-50-THUCTC (ours) & \bf 97.8\% \\
SC+ResNet-50 (ours) & 95.7\% \\
\hline
\end{tabular}
\end{center}
\caption{\label{table:fudan_results} Results of our Super Character (SC) method against other models on the Fudan dataset.}
\end{table}

\subsection{Analysis on the Impact of the Cut-length for Configuring Super Characters Image}
The cut-length determines how many characters in each generated Super Characters image. So it will impact if an input text needs clipping or padding, in order to have the same pixel size of each character and same length for all the text samples in the same dataset. The short cut-length may clip long sentences and cause information loss, which will decrease the sentiment classification accuracy. But increasing the cut-length of the text may prevent inadvertently clipping long sentences but also increases the number of blank spaces for short sentences. Thus it may impact the model and the sentiment analysis accuracy. The best setting for cut-length should be based on the dataset statistics. We have a study on different settings of cut-length using ResNet-50 on the Fudan corpus, and the results are given in Table \ref{cutlengthgraph}.　 For this Fudan data, the average sentence length is 530, and the median sentence length is 509. It shows that the setting of cut-length=784 is the best configuration for this dataset compared with other options. This indicates that setting the cut-length according to the median or average sentence length could be a good option.

\begin{table}[t!]
\begin{center}
\begin{tabular}{|c|c|c|c|c|c|c|c|c|c|c|c|c|}
\hline \bf Cut-length    &196  &256  &784  &1024  \\ \hline
\bf Accuracy(\%)  &93.45&93.3 &95.7 &89.35  \\ \hline
\end{tabular}
\end{center}
\caption{\label{cutlengthgraph} Cut-length Impact on Accuracy.}
\end{table}


\section{Conclusion and Future Work}
In this paper, we proposed the Super Characters method for sentiment classification. It converts text into images and then applies CNN models to classify the sentiment. The text features are automatically extracted by CNN models. We have tested our method on social media text contents from four different languages. The experimental results showed that our method consistently outperforms other methods for Chinese, English, Japanese, and Korean text contents for sentiment classification tasks. We also showed that pretrained Chinese text classification models on large datasets helps attain a higher accuracy for text classification on other Chinese datasets. 

For future work, we can apply various preprocessing techniques such as the elimination of common words and other methods to further increase the accuracy of this method, and fine-tune the cut length to analyze its impact on different data sets. And we also need to compare with other RNN methods on the same datasets.

\bibliography{emnlp2018}
\bibliographystyle{acl_natbib_nourl}

\end{document}